\documentclass[10pt, a4paper,twocolumn]{article}
\usepackage{ltc23}
\usepackage{graphicx}
\usepackage{url}

\usepackage{natbib}
\usepackage{multirow}
\title{CroSentiNews 2.0: A Sentence-Level News Sentiment Corpus}

\name{Gaurish Thakkar$^{\ast}$, Nives Mikelic
Preradovi{\'c}$^{\ast}$, Marko Tadi{\'c}$^{\ast}$} 
\address{ $^{\ast}$Faculty of Humanities and Social Sciences, University of Zagreb, Zagreb 10000 \\
                  gthakkar@m.ffzg.hr  \\
                  nmikelic@m.ffzg.hr \\ 
                  marko.tadic@ffzg.hr \\
                  }
\abstract{This article presents a sentence-level sentiment dataset for the Croatian news domain. In addition to the 3K annotated texts already present, our dataset contains 14.5K annotated sentence occurrences that have been tagged with 5 classes. We provide baseline scores in addition to the annotation process and inter-annotator agreement.}


\begin{document}

\maketitleabstract

\section{Introduction}
The objective of sentiment analysis (SA) is to categorize the orientation of the author's text 
\citep{DBLP:journals/coling/Cardie14}. The SA approaches have been analyzed for a variety of text types, including reviews \cite{kumar2019sentiment, 8320185}, news articles \citep{balahur-etal-2010-sentiment}, and social media \citep{rosenthal-etal-2017-semeval}. Prior work \citep{pang2008opinion,pang2002thumbs} in sentiment analysis primarily focused on document-level analysis. Since then, the research community has shifted its focus to fine-grained analysis \citep{narayanan2009sentiment,nguyen2015phrasernn,wang-etal-2016-attention,wang-etal-2017-tdparse}. Although there is a significant amount of research on high-resource languages, the same cannot be said for low-resource ones. In addition, the absence of annotated data limits the applicability of untested models because they cannot be evaluated on a test set. This study presents an enhanced version of the existing sentiment annotation dataset for Croatian sentences. 

SentiNews 1.0's sentiment corpus served as the basis for the initial dataset utilized to annotate SentiNews 2.0.
Croatian and Slovene language annotations can be found in the SentinNews 1.0 dataset.
Croatian is not annotated at the sentence or paragraph level in the SentiNews sentiment corpus, which is annotated at the document level. But it includes news on a range of subjects. This study discusses adding a sentence-level annotation layer to the SentiNews corpus; we think that such a dataset would be addition to the linguistic resources for low-resource languages like Croatian, in addition to providing a crucial building block for developing a sentiment classifier. The entire package will be made available to the public upon acceptance of this paper.

% The original dataset used for the annotation is the SentiNews sentiment corpus. The SentiNews sentiment corpus is entirely document annotated and contains neither paragraph nor sentence-level annotations for Croatian. However, it covers news from various topics. This paper addresses adding a sentence-level annotation to the SentiNews corpus; we believe that Such a dataset would be a valuable addition to the linguistic resources for low-resourced languages like Croatian, in addition, to providing a key component in building a sentiment classifier. All the resources listed along with guidelines and the associated code will be made public upon acceptance.
The following are the paper's main contributions:
(1) We provide CroSentiNews 2.0, a large dataset for sentiment analysis in Croatian news articles that has been manually annotated.
(2) We perform an extensive evaluation of the dataset using a pre-trained transformer model. 
To improve individual classification performance, we employ the multi-task architecture \citep{thakkar2021multi}, which makes use of a multi-level dataset.

% The main contributions of the paper are as follows: (1) We introduce CroSentiNews 2.0, a large manually annotated dataset for sentiment analysis in Croatian news articles. (2) We perform an extensive evaluation of the dataset using a pre-trained transformer model. We use the multi-task architecture \citep{thakkar2021multi} which uses a multi-level dataset to improve individual classification performance. Apart from comparing the classification scores, the current work introduces a curated dataset with a large number of annotations, annotated by multiple annotators.

% The paper is structured as follows. In Section 2, we review the related work in sentiment analysis and data annotation. In Section 3, we describe our overall annotation strategy and summarise the dataset. In Section 4 We describe the details of the experiments. We report the results in Section 5. Before concluding we provide a summary and future work in Section 6.
The paper is structured as follows: In Section 2, we review the related work in sentiment analysis and data annotation. In Section 3, we describe our overall annotation strategy and summarize the dataset. In Section 4, we describe the details of the experiments. We report the results in Section 5. Before concluding, we provide a summary and future work in Section 6.

\section{Related work}
% Below we note down the related work on existing datasets and sentiment modelling approaches in Croatian.
% \subsection{Datasets}
% The contributions towards sentiment classification in a language have relied on the availability of the dataset. 
% For Croatian sentiment analysis,  Among the few datasets that exist for sentiment analysis
% The earliest attempt at Csentiment dataset creation relied on
Below, we note the related work on existing datasets and sentiment modelling approaches in Croatian.

% One of the earliest attempts at Croatian sentiment classification is the contribution of , in which authors designed a rule-based model for the automatic detection of general sentiment and polarity phrases in Croatian text from the finance domain.
\citet{agic-etal-2010-towards} made one of the first attempts at Croatian sentiment classification, developing a rule-based model for the automatic detection of general sentiment and polarity phrases in Croatian text from the finance domain.
Using the text from the game reviews, \citet{rotim-snajder-2017-comparison} compiled a dataset with sentiment marked with three-class label and compared the performance of the Support Vector Machine (SVM) classification algorithm on two additional Twitter-derived short-text Croatian datasets.

\citet{10.1007/978-981-16-1781-2_35} utilized a sentiment lexicon in Croatian to annotate a training dataset for  the classifier.
A dataset for stance, claim, and sentiment for Croatian was presented by \citet{bosnjak-karan-2019-data}.
 % analyzed sentiments and emotions in crisis communication in the news related to the COVID-19 pandemic using topic modelling and word-emotion lexicons.
Using topic modelling and word-emotion lexicons, \citet{pandur2020topic} studied sentiments and emotions in crisis communication in the news connected to the COVID-19 epidemic.
\citet{app10175993} compiled a sentiment-annotated dataset for Croatian and conducted experiments in zero-shot settings.
% \citet{Robnik-Šikonja_Reba_Mozetič_2021,martina2021article} investigated state-of-the-art approaches to the cross-lingual transfer of Twitter sentiment prediction models for 13 European languages, including Croatian, using the multi-lingual dataset \citep{11356/1054}. 
Using a multilingual Twitter sentiment dataset \citep{10.1371/journal.pone.0155036}, \citet{Robnik-sikonja_Reba_Mozetic_2021} and, \citet{martina2021article} evaluated state-of-the-art approaches to the cross-lingual transfer of sentiment prediction models for 13 European languages, including Croatian. 

Earlier attempts at SA typically depended on rules and lexicons \citep{Hutto_Gilbert_2014,baccianella-etal-2010-sentiwordnet}. A majority of the recently proposed work \citep{sun-etal-2019-utilizing} relies on fine-tuning the BERT\citep{devlin-etal-2019-bert} for downstream tasks. Continual pre-training \citep{gururangan-etal-2020-dont} using unlabelled data has also yielded great results for domain-specific sentiment classification.
% \subsection{Modeling}

\section{Annotations}
% For creating the Cro-entiNews 2.0, we used the Croatian Sentiment Dataset \citep{app10175993} created based on the guidelines from \citet{10.1007/978-3-319-26227-7_73}. The dataset is based on news articles from the 24sata website which is the leading media company in Croatia. Apart from the daily news, the news text spans topics like automotive news, health, culinary content, and lifestyle advice. The dataset statistics are:
CroSentiNews 2.0 is built using the Croatian Sentiment Dataset \citep{app10175993}, which was compiled in accordance with the guidelines from \citet{10.1007/978-3-319-26227-7_73}.
The collection consists of news articles from the website of 24sata, the major media organization in Croatia.
The news text includes topics such as automotive news, health, culinary content, and lifestyle advice, in addition to daily news. The dataset's statistics are as follows:

\begin{itemize}
    \item 2,025 documents;
    \item 12,032 paragraphs;
    \item 25,074 sentences.
\end{itemize}
The documents were initially tagged on a five-point Likert \citep{likert1932technique} scale (very negative, negative, neutral, positive, very-positive) and subsequently projected onto three-class labels (negative, neutral, positive). Contrary to its Slovenian counterpart, the Croatian SentiNews 1.0 contains documents with sentiment labels but lacks paragraph- and sentence-level annotations. Due to a large number of sentences, we opted to leave the annotation of sentiment at the paragraph level for future work. 
To prepare the sentences for annotations, we performed sentence tokenization on the whole document. All the sentences were divided into nine groups such that no annotator received a part of the document. Each group consisted of two or more annotators, with an average of approximately 2,114 sentences. Before presenting the sentences to the annotators, the text was pre-annotated using an existing sentiment classifier \citep{thakkar2021multi}. Thus reducing the sentiment annotation problem to simply label correction if the labels didn't match the text. All the annotators were native speakers of Croatian enrolled in undergraduate courses in linguistics. The total numbers of annotators at the beginning of the crowdsourcing process were 20 but only 16 completed the whole task. 

\subsection{Annotation guidelines}
Following the guidelines outlined in \citep{mohammad-2016-practical} and \citep{app10175993}, we produced annotation guidelines detailing the complete annotation technique. Annotators were presented with five labels of sentiment:  1—negative, 2—neutral, 3—positive, 4—mixed, and 5—other/sarcasm. Each label was illustrated with an example. The overall annotation was conducted using the INCEpTION \citep{tubiblio106270} tool and a detailed user manual was provided. No user was allowed to view documents from groups other than the ones assigned to him. The documents were marked as complete using the locking mechanism that freezes the annotation for a document. Eventually, statistics of complete documents were derived for the locked documents. We measured the reliability of the agreement using the Fleiss Kappa \citep{fleiss1971mns} score across multiple groups. The scores range from moderate (0.41-0.60) to substantial (0.61-0.80) levels of annotator agreement.

\subsection{Data statistics}
Out of all the sentences, only 19k instances were tagged by at least one annotator. Out of which only 14.5k had a total agreement. We filtered out the sentences not having any labels. Thus, the corpus consists of 1,988 unique non-empty documents and 14,5k sentiment-labelled phrases. Moreover, there were 428 cases of mixed language and 73 instances of sarcasm, but we did not include these in our studies. Instances marked by two annotators who did not agree with the label were likewise eliminated. The final label for a sentence was determined by the majority vote of the sentence's annotators. The distribution of labels for the CroSentiNews 2.0 and SentiNews dataset is depicted in Table \ref{table:1}. The SentiNews 1.0 dataset utilized in our tests is displayed in table \ref{table:4} . 
% Thus, the corpus consists of 1988 unique non-empty documents and 14.5k sentences that are tagged for sentiment labels. In addition, there were 428 mixed and 73  sarcasm instances but we did not use them in our experiments. We also filtered out instances that were tagged by even annotators and did not agree upon the label. The final label was selected by the majority vote of the annotators for a sentence. Table  shows the overall distribution of labels for the Croatian and Slovene SentiNews datasets.

\begin{table*}[!h]
\centering
  \begin{tabular}{llclll}
    \hline
    Language  & Level   & Total Instances & Positive & Negative  & Neutral\\
    \hline
    Croatian  & Document      & 1,988 & 321& 450& 1,217\\
    {}        & Sentence       & 14,570 & 3,265 & 3,353  & 3,265 \\ \hline
    
    \hline
  \end{tabular}
\caption{Distribution of CroSentiNews 2.0 dataset \label{table:1}}
\end{table*}

\begin{table*}[!h]
\centering
  \begin{tabular}{llclll}
    \hline
    Language  & Level   & Total Instances & Positive & Negative  & Neutral\\
    \hline
   
    Slovene & Document       & 10,417 & 1,665 &3,337 &5,418\\
    {}        & Paragraph      & 86,803 &14,270&23,265 &49,268\\
    {}        & Sentence       & 161,291 &26,679 &44,014 &90,598\\ \hline
      
    \hline
  \end{tabular}
\caption{Distribution of SentiNews 1.0 dataset \label{table:3}}
\end{table*}

\begin{table*}[ht!]
\centering
\begin{tabular}{|c|c|cccc|}
\hline
\begin{tabular}[c]{@{}c@{}}\# of \\ classification\\  heads\end{tabular} & Train set & \multicolumn{4}{c|}{Test set} \\ \hline
 & \multirow{2}{*}{} & \multicolumn{2}{c|}{Croatian} & \multicolumn{2}{c|}{Slovene} \\ \cline{1-1} \cline{3-6} 
 &  & \multicolumn{1}{c|}{Document} & \multicolumn{1}{c|}{Sentence} & \multicolumn{1}{c|}{Document} & Sentence \\ \hline 
3 head \citep{thakkar2021multi} & SL MTL & \multicolumn{1}{c|}{50.07} & \multicolumn{1}{c|}{} & \multicolumn{1}{c|}{\textbf{74.86}} & \textbf{69.40} \\ 
3 head \citep{thakkar2021multi} & SL+HR MTL & \multicolumn{1}{c|}{63.86} & \multicolumn{1}{c|}{} & \multicolumn{1}{c|}{74.21} & 69.21 \\  \hline
 &  & \multicolumn{1}{c|}{} & \multicolumn{1}{c|}{} & \multicolumn{1}{c|}{} &  \\  \hline
1 head & HR\_Doc - STL & \multicolumn{1}{c|}{59.24} & \multicolumn{1}{c|}{60.67} & \multicolumn{1}{c|}{49.54} & 46.87 \\ 
1 head & HR\_Sent - STL & \multicolumn{1}{c|}{55.00} & \multicolumn{1}{c|}{77.72} & \multicolumn{1}{c|}{54.99} & 56.29 \\  
1 head & HR\_Sent+SL\_Sent & \multicolumn{1}{c|}{57.71} & \multicolumn{1}{c|}{72.24} & \multicolumn{1}{c|}{63.15} & 64.28 \\  
2 head & HR\_Doc+SL\_Sent - MTL & \multicolumn{1}{c|}{64.46} & \multicolumn{1}{c|}{65.15} & \multicolumn{1}{c|}{68.22} & 67.77 \\  
2 head & HR\_Doc+HR\_Sent - MTL & \multicolumn{1}{c|}{\textbf{65.60}} & \multicolumn{1}{c|}{\textbf{79.65}} & \multicolumn{1}{c|}{55.83} & 56.97 \\  \hline
\end{tabular}
\caption{\label{table:2} Results of experiments. The first two rows contain values from the original paper.}
\end{table*}

\section{Experiments}
\begin{figure}[t!]
\begin{center}
    \includegraphics[width=\linewidth]{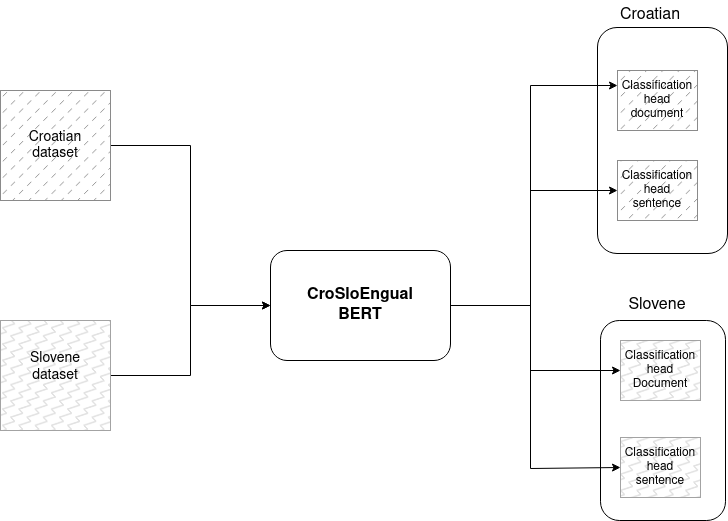} %TODO replace with following for black and white LTC23_latex_style/LTC23.draio-bw.png
    \caption{Multitask training setup using Croatian and Slovene datasets.}
\end{center}
\label{fig:fig1}
\end{figure}

We employed two experimental setups to perform bench-marking on the newly compiled dataset, namely single-task fine-tuning and multitask fine-tuning. In a single-task fine-tuning setup \citep{devlin-etal-2019-bert}, a classification head is added on top of the pre-trained model and all the parameters are jointly trained using the dataset from the downstream task. We modified the model presented in \citet{thakkar2021multi} for the multitask learning (MTL) setup by eliminating the paragraph classification head, as we lacked paragraph-level annotations.
Still employing multi-level labels, the approach is restricted to sentence- and document-level labels.
The models were trained using the following combinations of datasets. 
% In the MTL setup, we adapted the model from \citep{thakkar2021multi} by simply removing the paragraph classification head as we did not have paragraph-level annotations. The model still used the multi-level labels but is confined to using sentence and document-level labels. 
% The models were trained in the following dataset combinations.

% \subsection{Single-task fine-tuning}

% \subsection{Multi-task fine-tuning}
\begin{enumerate}
    \item HR\_Document\_STL—The model consists of a single classification head that is trained end-to-end using a Croatian document-level supervised dataset.
    \item HR\_Sentence\_STL—Similar to the previous scenario, except that the model is trained with sentence-level annotations in only Croatian. 
    \item HR\_Sentence+SL\_Sentence—The model consists of a single classification head trained on sentences from datasets in Croatian and Slovene.
    \item HR\_Document+SL\_Sentence\_MTL-Based on the MTL model from \cite{thakkar2021multi} with no paragraph-level classification head. The model consists of two classification heads, one for document-level and another for sentence-level classification. The dataset used for training comprises Croatian documents and Slovene sentences. 
    \item HR\_Document+HR\_Sentence\_MTL—Similar to the preceding scenario, but limited to Croatian and without dataset inter-mixing.
\end{enumerate}

\subsection{Model training}
% In all the experiments we used the CroSloEngual BERT \citep{ulcar-robnik2020finest} as the pre-trained language model. All the datasets were divided into 80-20 train-test splits. For the single-task training, we performed a population-based hyper-parameter search on 10\% of the training set and found the following values for the Croatian document and sentence classification: learning rate= (2.8e-05, 3.9e-05), weight decay= (0.15,0.28), batch size=16. Early stopping was performed when there was no evident improvement in the evaluation loss. For the MTL, we used a minibatch size of 32, a learning rate of 2e-05 and a hidden state dropout of 0.3. It took 5 epochs for the system to converge. 
We utilized the CroSloEngual BERT \citep{ulcar-robnik2020finest} as the pre-trained language model in all trials.
All datasets were divided into 80–20 splits for training and testing.
For the single-task training, a population-based hyperparameter search on 10\% of the training set yielded the values for the classification of Croatian documents and sentences depicted in Table \ref{table:4}.  
% learning rate= (2.8e-05, 3.9e-05), weight decay= (0.15,0.28), and batch size=16.
The trial was terminated early when there was no discernible improvement in the evaluation loss.
We utilized a minibatch size of 32, a learning rate of 2e-05, and a hidden state dropout of 0.3 for the MTL.
Five epochs were required for the system to converge. 

% \usepackage{multirow}
\begin{table}[ht]
\begin{tabular}{|l|l|l|}
\hline
                                           & Parameters        & Values \\ \hline
\multicolumn{1}{|c|}{\multirow{4}{*}{STL}} & learning rate\_document    & 2.8e-05         \\ %\cline{2-3} 
\multicolumn{1}{|c|}{}                     & learning rate\_sentence    & 3.9e-05         \\ %\cline{2-3} 
\multicolumn{1}{|c|}{}                     & weight decay\_document     & 0.15            \\ %\cline{2-3} 
\multicolumn{1}{|c|}{}                     & batch size                 & 16              \\ \hline
\multirow{5}{*}{MTL}                       & weight decay STL\_sentence & 0.28            \\ %\cline{2-3} 
                                           & learning rate              & 2e-05           \\ %\cline{2-3} 
                                           & weight decay               & 0.0             \\ %\cline{2-3} 
                                           & batch size                 & 32              \\ %\cline{2-3} 
                                           & hidden state dropout       & 0.3             \\ \hline
\end{tabular}
\caption{\label{table:4} Hyperparameters and their values.}
\end{table}

\subsection{Results}
% As the datasets are skewed choose macro-F1 as an evaluation metric. Table \ref{table:2} shows the macro-F1 scores for different combinations of datasets. For the sake of comparison, we also report the macro-F1 on the Slovene document and sentence-level classification. For each dataset, we ran the experiment with 5 seeds and reported the mean.
% The best performing method for document-level and sentence-level classification used HR-document and HR-sentence-level annotations and were trained in MTL fashion. We believe improvements are significant due to the model's ability to learn from the same text and same language but for different tasks.
% In almost all the cases, we observed that the Slovene classification scores did not improve when combined with Croatian training data. On the other hand, training Croatian (documents) and Slovene (sentence) combined datasets for different levels of annotations using MTL architecture did improve the scores for the Croatian document classification task. This hints at using data from typologically similar language in an MTL setup could benefit when the source language has a lower number of instances ($<$ 2000 for Croatian). But in the case where a large number of instances are available ($>$ 14.5k) combining with Slovene decreases performance for Croatian sentence-level classification. Owing to a large number of annotated instances, the BERT-based sentence-level classification model did perform the second best in terms of scores. 
% We analysed the confusion matrix of the best-performing model and found the 
As the datasets are skewed, macro-F1 is used as the metric of evaluation.
Table 2 displays the macro-F1 scores for various combinations of datasets.
For the sake of comparison, we also offer the macro-F1 classification of Slovene documents and sentences.
We repeated the experiment with 5 seeds for each dataset and reported the mean.

The most effective strategy for document-level and sentence-level categorization employed HR\_document and HR\_sentence-level annotations that were trained in an MTL-like fashion.
We believe the model's ability to learn from the same text and language, but for different tasks, has led to considerable advances.

We discovered that the Slovene classification scores did not improve when coupled with Croatian training data in almost every instance. On the other hand, training Croatian (documents) and Slovene (sentences) mixed datasets with different degrees of annotations using MTL architecture enhanced the performance of the Croatian document classification task. This suggests that using data from a typologically similar language in an MTL setup could be advantageous when the source language contains fewer instances ($<$2,000 or less for Croatian). 
% When a large number of instances are available, wever ($>$14.5k), combining with Slovene affects the performance of sentence-level categorization in Croatian. 
In terms of scores, the BERT-based sentence-level classification model performed second-best due to a large number of annotated examples. 

% During the training with the combination of Slovene and Croatian sentence-level instances, we found that this combination took longer time to converge than any combination. This could be due to the size of the combined dataset but it did not provide any significant improvements over the BERT-based baseline.
During training using the mix of Slovene and Croatian sentence-level examples, we discovered that this combination required the most time to converge.
This may be related to the magnitude of the combined dataset, but it did not result in any notable gains over the BERT-based baseline.

% We conducted a hyperparameter optimization on parameters like learning rate, weight decay, batch size and number of epoch. 

% We  found  out  that  the  best  result were obtained with a minibatch size of 25, Tree-LSTM hid-den state and cell state size of 300, learning rate of 0.05, weight  decay  rate  of  0.0001  and  L2  regularization  rate of 0.0001.   
% It takes 3-5 epochs for the system to converge.

% performs competitively with state-of-the-art methods. The best performing method concatenates the token representations from the top four hidden layers of the pre-trained Transformer, which is only
% 0.3 F1 behind fine-tuning the entire model. This demonstrates that BERT is effective for both fine-tuning and feature-based approaches
% \subsection{Discussions}

% but we hypothesize that this combination also acts as regularising factor not allowing the model to overfit on a single subset of the dataset.

% Our initial experiments pointed out the fact that combining data from Croatian and English in STL did not provide any significant improvement over the HR+SL combination.

\section{Conclusion}
% In this paper, we introduced Cro-SentiNews 2.0 dataset for Croatian. The dataset is based on an existing dataset having document-level sentiment annotations but is enhanced with a layer of sentence-level annotation. We have described our annotation procedure and presented the statistics of the final dataset. Furthermore, we have performed experiments using single-task and multi-task frameworks and reported the performance scores of the same. In future, we want to extend the work with other Slavic and non-Slavic language datasets using MTL setup.  
In this paper, we introduced the CroSentiNews 2.0 dataset for Croatian. The dataset is derived from an existing dataset with document-level sentiment annotations and is augmented with a sentence-level annotation layer. We have discussed our annotation technique and presented the statistics of the final dataset. In addition, we have conducted studies employing single-task and multi-task frameworks and published their respective performance results. In the future, we want to incorporate other Slavic and non-Slavic language datasets utilizing MTL configuration. 

\section{Acknowledgements}
The work presented in this paper has received funding from the European Union’s Horizon 2020 research and innovation program under the Marie Skłodowska-Curie grant agreement no. 812997 and under the name CLEOPATRA (Cross-lingual Event-centric Open Analytics Research Academy).
% \section{Limitations}
% A key potential limitation of the current method is the 

\section{Ethical consideration}
We work on a sentiment classification task, which is a standard NLP problem. Based on our experiments, we do not see any major ethical concerns with our work. We would like to note that, depending on their source, news articles and their annotators can be politically biased.

\bibliographystyle{ltc23}
\bibliography{xample23}

\end{document}